\documentclass[conference]{IEEEtran}
\IEEEoverridecommandlockouts

\usepackage[super]{nth}
\usepackage{graphicx}
\usepackage{textcomp}
\usepackage[table, x11names]{xcolor}
\usepackage{multirow}
\usepackage{placeins}
\usepackage{enumitem}
\usepackage{bm}
\usepackage{booktabs}
\usepackage{caption}
\usepackage{hyperref}
\usepackage{footmisc}

\usepackage{amsmath,amssymb,amsfonts}
\usepackage{mathtools}

\usepackage{algorithmic}
\usepackage{xfrac}

\usepackage{graphicx}
\usepackage{adjustbox}
\usepackage{subfig}
\usepackage{float}
\usepackage{tikz}
\usepackage{wrapfig}
\usetikzlibrary{positioning,chains}

\usepackage{cite}

\graphicspath{ {plots/} }

\def\BibTeX{{\rm B\kern-.05em{\sc i\kern-.025em b}\kern-.08em
    T\kern-.1667em\lower.7ex\hbox{E}\kern-.125emX}}
    
\begin{document}

\title{From Theory to Behaviour: \\
Towards a General Model of Engagement
\thanks{This work was supported by the EPSRC Centre for Doctoral Training in Intelligent Games \& Games Intelligence (IGGI) [EP/L015846/1] and the Digital Creativity Labs (digitalcreativity.ac.uk), jointly funded by EPSRC/AHRC/Innovate UK under grant no. EP/M023265/1.}
}

\author{
\IEEEauthorblockN{
    V. Bonometti\dag\ddag, 
    C. Ringer\dag, 
    M. Ruiz\ddag, 
    A.R. Wade\S, 
    A. Drachen\dag
}
\IEEEauthorblockA{
    \dag Department of Computer Science, University of York, York, UK YO10 5DD\\
    \ddag Square Enix Limited, London, UK SE1 8NW\\
    \S Department of Psychology, University of York, York, UK YO10 5DD \\
Email: \{vb690, cr1116, anders.drachen, alex.wade\}@york.ac.uk; analyticsl@eu.square-enix.com}
}
\maketitle



\begin{abstract}
Engagement is a fuzzy concept. In the present work we operationalize engagement mechanistically by linking it directly to human behaviour and show that the construct of engagement can be used for shaping and interpreting data-driven methods. First we outline a formal framework for engagement modelling. Second we expanded on our previous work on theory-inspired data-driven approaches to better model the engagement process by proposing a new modelling technique, the Melchoir Model. Third, we illustrate how, through model comparison and inspection, we can link machine-learned models and underlying theoretical frameworks. Finally we discuss our results in light of a theory-driven hypothesis and  highlight potential application of our work in industry. \footnote{Extensive ancillary information and results can be found at https://github.com/vb690/modelling\_engagement\_ammount/wiki.}
\end{abstract}



\section{INTRODUCTION}
Various attempts have been made in prior literature across several domains to describe and analyse the engagement process in a video-game context \cite{boyle2012engagement}. Engagement is an extremely useful construct because it allows us to understand  how a user (e.g. a player) is interacting with a specific object or activity (i.e. a video game) and how this interaction can evolve over time. Here, we build upon our previous work \cite{bonometti2019modelling} modeling engagement and strengthen its link with human behaviour. We expand our previous theoretical framework and use it to refine our modelling approach. In doing so we generate a set of theory-driven hypothesis that we investigate through model comparison and inspection. This was achieved by carrying out a series of three analyses over a 3.2 millions-entries cross-game data-set. Our goal was not just to evaluate the effectiveness of our new approach but also to ask how the assumptions generated by the theoretical framework compare to those learned by our modelling approach. In the final section we briefly discuss a series of practical application that our current approach can have in industry settings, ranging from estimation of engagement evolution and identification of engagement profiles. As far as we know this is one of the first works explicitly translating qualitative theories of engagement into machine-learned models. We argue that our refined theoretical framework allows for a re-framing of engagement in behavioural terms while also allowing us to formulate and test more precise hypotheses. The modelling approach we propose here extends and refines our previous model  - maintaining the same advantages while also improving on some of its limitations. In particular, we explicitly consider the contribution of environmental variables, and we fully exploit the temporal nature of the engagement process allowing the model to work continuously with an arbitrarily long sequence of input features. Finally, the model is able to estimate a larger range of target metrics when compared with its predecessor. This work also presents a methodology for inspecting, analyzing and interpreting the user representation learned by the model, providing a generalizable way of producing understandable engagement profiles. 


\section{STATE OF THE ART AND CONTRIBUTION}
In this section we give a general overview on the state of the art in engagement modelling. Due to space constraints and the substantial literature on the matter we will focus on a restricted set of representative works in the area of large-scale behavioural modelling of engagement. The work on engagement modelling comes, generally, in two forms: estimation and profiling of in-game behaviour \cite{el2016game}. The estimation of in-game behaviour is usually formulated as a supervised machine-learning or more general statistical modelling problem\cite{el2016game}. Despite the literature on the topic often presenting compelling solutions for practical problems, it tends to follow a black box approach: a machine-learned solution is generated for solving a specific task but no attempts are made to inspect or interpret the model \cite{lee2018game, liu2019micro, del2020time, kristensen2019combining}. Moreover, when these attempts are made, the lack of a solid and predefined theoretical framework tends to lead to post-hoc interpretations which are sometimes difficult to verify or relate with actual human behaviour \cite{drachen2016rapid, del2019profiling}. When trying to estimate engagement profiles, the approach widely used in the literature is to adopt some form of unsupervised learning technique for individuating patterns of interaction with various in-game features \cite{el2016game, del2019profiling}. This however is usually done considering an unconstrained set of game-specific metrics. As a result, \textit{a-posteriori} justifications for the characteristics of the individuated profiles are provided \cite{drachen2012guns, makarovych2018like, drachen2009player}, which, without an overarching explanatory theoretical framework, appear to be be very context-specific and difficult to interpret. What we see in the literature is that attempts are made to model a single behavioural manifestation of engagement rather than the construct in its entirety. A noticeable exception in this regard is the recent work by Reguera et al. \cite{reguera2020quantifying}, who adopt a complete data-driven approach managed to derive a general law for describing and quantifying the engagement process, similarly to what Bauckhage at all. did in \cite{bauckhage2012players}. However, neither group interpret their findings through the lens of existing human behaviour theory. We believe that a holistic model of engagement can be generated, constraining the great flexibility provided by data-driven approaches by employing solid and well established theoretical priors \cite{yannakakis2013player}. To do so, an \textit{a-priori} theoretical framework which is guaranteed to generalise to different situations should be defined. Such a framework should clearly state what are the observable and measurable indicators of engagement and how they are expected to vary in relations with the construct's dynamics. In doing so the findings emerging from data-driven approaches can be compared with what the theoretical framework prescribes. 


\section{ENGAGEMENT AS A BEHAVIOURAL PROCESS}
Although various attempts have been made to describe the construct of engagement, prior literature struggles to provide a formal definition \cite{boyle2012engagement}. In particular, we observed a lack of mechanistic explanations and parallelisms with human behaviour in favour of more holistic and phenomenological descriptions of the concept \cite{boyle2012engagement}. Although these phenomenological descriptions do provide qualitative insights, mechanistic explanations are necessary to generate and verify hypotheses as well as producing real-world applications for engagement modelling. For this reason, we extend the theoretical framework adopted in our previous work \cite{bonometti2019modelling} to draw stronger connections between engagement and human behaviour. This allowed us to not only make more informed decisions when designing a strategy for modelling engagement in behavioural terms but also to use this approach for verifying a set of theory driven hypotheses.

\subsection{The Engagement Process Model}
\begin{figure}[h]
    \begin{center}
    
    \begin{adjustbox}{width=0.75\columnwidth}
    
        \tikzset {_07jciofbh/.code = {\pgfsetadditionalshadetransform{ \pgftransformshift{\pgfpoint{0 bp } { 0 bp }  }  \pgftransformrotate{0 }  \pgftransformscale{2 }  }}}
        \pgfdeclarehorizontalshading{_79lkkjg9h}{150bp}{rgb(0bp)=(0.97,0.02,0.02);
        rgb(37.5bp)=(0.97,0.02,0.02);
        rgb(62.5bp)=(0.07,0.04,0.92);
        rgb(100bp)=(0.07,0.04,0.92)}
        \tikzset{_m7pe8g662/.code = {\pgfsetadditionalshadetransform{\pgftransformshift{\pgfpoint{0 bp } { 0 bp }  }  \pgftransformrotate{0 }  \pgftransformscale{2 } }}}
        \pgfdeclarehorizontalshading{_d5g9qtq13} {150bp} {color(0bp)=(transparent!75);
        color(37.5bp)=(transparent!75);
        color(62.5bp)=(transparent!75);
        color(100bp)=(transparent!75) } 
        \pgfdeclarefading{_nvxzylyb8}{\tikz \fill[shading=_d5g9qtq13,_m7pe8g662] (0,0) rectangle (50bp,50bp); } 
        \tikzset{every picture/.style={line width=0.75pt}} 
        
        \begin{tikzpicture}[x=0.75pt,y=0.75pt,yscale=-1,xscale=1]
        
        \draw  [fill={rgb, 255:red, 247; green, 4; blue, 4 }  ,fill opacity=0.5 ] (179.8,67.66) .. controls (179.8,63.17) and (183.44,59.52) .. (187.93,59.52) -- (272.24,59.52) .. controls (276.73,59.52) and (280.37,63.17) .. (280.37,67.66) -- (280.37,92.06) .. controls (280.37,96.55) and (276.73,100.19) .. (272.24,100.19) -- (187.93,100.19) .. controls (183.44,100.19) and (179.8,96.55) .. (179.8,92.06) -- cycle ;
        \draw    (229.93,99.73) -- (229.87,114.14) ;
        \draw [shift={(229.86,117.14)}, rotate = 270.25] [fill={rgb, 255:red, 0; green, 0; blue, 0 }  ][line width=0.08]  [draw opacity=0] (10.72,-5.15) -- (0,0) -- (10.72,5.15) -- (7.12,0) -- cycle    ;
        \draw  [fill={rgb, 255:red, 247; green, 4; blue, 4 }  ,fill opacity=0.25 ] (179.51,127.56) .. controls (179.51,123.07) and (183.16,119.43) .. (187.65,119.43) -- (271.95,119.43) .. controls (276.44,119.43) and (280.09,123.07) .. (280.09,127.56) -- (280.09,151.96) .. controls (280.09,156.45) and (276.44,160.1) .. (271.95,160.1) -- (187.65,160.1) .. controls (183.16,160.1) and (179.51,156.45) .. (179.51,151.96) -- cycle ;
        \draw    (280.09,140.4) -- (315.09,140.4) ;
        \draw [shift={(318.09,140.4)}, rotate = 180] [fill={rgb, 255:red, 0; green, 0; blue, 0 }  ][line width=0.08]  [draw opacity=0] (10.72,-5.15) -- (0,0) -- (10.72,5.15) -- (7.12,0) -- cycle    ;
        \draw  [fill={rgb, 255:red, 18; green, 11; blue, 234 }  ,fill opacity=0.25 ] (320.37,128.79) .. controls (320.37,124.43) and (323.91,120.89) .. (328.28,120.89) -- (412.18,120.89) .. controls (416.55,120.89) and (420.09,124.43) .. (420.09,128.79) -- (420.09,152.51) .. controls (420.09,156.88) and (416.55,160.41) .. (412.18,160.41) -- (328.28,160.41) .. controls (323.91,160.41) and (320.37,156.88) .. (320.37,152.51) -- cycle ;
        \path  [shading=_79lkkjg9h,_07jciofbh,path fading= _nvxzylyb8 ,fading transform={xshift=2}] (249.51,187.66) .. controls (249.51,183.2) and (253.13,179.59) .. (257.58,179.59) -- (341.73,179.59) .. controls (346.19,179.59) and (349.8,183.2) .. (349.8,187.66) -- (349.8,211.85) .. controls (349.8,216.31) and (346.19,219.92) .. (341.73,219.92) -- (257.58,219.92) .. controls (253.13,219.92) and (249.51,216.31) .. (249.51,211.85) -- cycle ; 
         \draw  [dash pattern={on 4.5pt off 4.5pt}] (249.51,187.66) .. controls (249.51,183.2) and (253.13,179.59) .. (257.58,179.59) -- (341.73,179.59) .. controls (346.19,179.59) and (349.8,183.2) .. (349.8,187.66) -- (349.8,211.85) .. controls (349.8,216.31) and (346.19,219.92) .. (341.73,219.92) -- (257.58,219.92) .. controls (253.13,219.92) and (249.51,216.31) .. (249.51,211.85) -- cycle ; 
        
        \draw    (370.33,120.89) -- (370.37,104.51) ;
        \draw [shift={(370.38,101.51)}, rotate = 450.13] [fill={rgb, 255:red, 0; green, 0; blue, 0 }  ][line width=0.08]  [draw opacity=0] (10.72,-5.15) -- (0,0) -- (10.72,5.15) -- (7.12,0) -- cycle    ;
        \draw  [fill={rgb, 255:red, 18; green, 11; blue, 234 }  ,fill opacity=0.5 ] (320.37,67.43) .. controls (320.37,62.98) and (323.98,59.37) .. (328.44,59.37) -- (412.59,59.37) .. controls (417.05,59.37) and (420.66,62.98) .. (420.66,67.43) -- (420.66,91.63) .. controls (420.66,96.09) and (417.05,99.7) .. (412.59,99.7) -- (328.44,99.7) .. controls (323.98,99.7) and (320.37,96.09) .. (320.37,91.63) -- cycle ;
        \draw  [dash pattern={on 4.5pt off 4.5pt}]  (230.57,165.7) .. controls (230.32,200.22) and (230.67,199.83) .. (249.51,199.83) ;
        \draw [shift={(230.6,162.4)}, rotate = 90.44] [fill={rgb, 255:red, 0; green, 0; blue, 0 }  ][line width=0.08]  [draw opacity=0] (10.72,-5.15) -- (0,0) -- (10.72,5.15) -- (7.12,0) -- cycle    ;
        \draw  [dash pattern={on 4.5pt off 4.5pt}]  (370.14,162.57) .. controls (369.57,200.29) and (370.09,200.11) .. (349.8,200.11) ;
        \draw   (160.53,39.97) -- (439.53,39.97) -- (439.53,240.3) -- (160.53,240.3) -- cycle ;
        
        \draw (230.83,79.86) node  [font=\fontsize{0.67em}{0.8em}\selectfont] [align=center] {
        {\large Point of}\\{\large Engagement}
        };
        \draw (230.5,139.76) node  [font=\fontsize{0.67em}{0.8em}\selectfont] [align=center] {
        {\large Sustained}\\{\large Engagement}
        };
        \draw (370.01,140.25) node  [font=\fontsize{0.67em}{0.8em}\selectfont] [align=center] {
        {\large Dis}\\{\large Engagement}
        };
        \draw (369.91,79.53) node  [font=\fontsize{0.67em}{0.8em}\selectfont] [align=center] {
        {\large Extinction}
        };
        \draw (300.67,199.76) node  [font=\fontsize{0.67em}{0.8em}\selectfont] [align=center] {
        {\large Re}\\{\large Engagement}
        };
        \draw (212.33,27) node   [align=left] {{\large \textit{Env}}};
        \draw (166.53,42.97) node [anchor=north west][inner sep=0.75pt]   [align=left] {{\large 1}};
        \draw (166.53,102.97) node [anchor=north west][inner sep=0.75pt]   [align=left] {{\large 2}};
        \draw (421.53,102.97) node [anchor=north west][inner sep=0.75pt]   [align=left] {{\large 3}};
        \draw (236.53,215.97) node [anchor=north west][inner sep=0.75pt]   [align=left] {{\large 4}};
        \draw (421.53,42.97) node [anchor=north west][inner sep=0.75pt]   [align=left] {{\large 5}};

        \end{tikzpicture}
        
    \end{adjustbox}

    \end{center}
\caption{\textbf{The Engagement Process Model.} Solid and dashed lines represent compulsory and  optional paths.}
\label{fig: epm}
\end{figure}
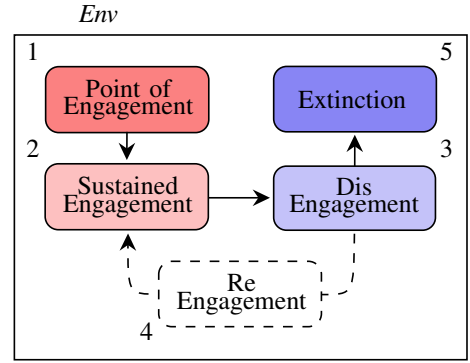

When looking at the various formulation of engagement \cite{boyle2012engagement} a common denominator seems to emerge: from a behavioural point of view engagement can be represented with the amount, duration and frequency of interactions between an individual $I$ and an object $O$. The Engagement Process Model proposed by O'Brien and Toms \cite{o2008user} perfectly summarizes this, describing these interactions in terms of a dynamic system. In their system, the ability of $O$ to provide rewarding experiences to $I$, in conjunction with environmental factors $Env$, controls cycles of interaction between $I$ and $O$. Figure \ref{fig: epm} represents the process by which $I$ engages with $O$ inside $Env$. If $I$ has an \textit{a-priori} belief that $O$ is able to provide rewarding experiences they will direct their attention towards it (1). If $Env$ does not pose any constraints, $I$ will interact with $O$ for as long as $O$ is able to provide rewarding experiences (2). 

However, if $O$ fails to provide these experiences, or constraints from $Env$ emerge, $I$ will gradually begin to disengage from $O$ (3). At this point $I$ can either enter a cycle of re-engagement and disengagement (4) or reach an inevitable state of complete withdrawal from $O$ (5).

\subsection{Incentive Salience Attribution}
The Engagement Process Model provides a good high-level description of the engagement process but fails to give a clear explanation of the dynamics controlling the system or explicitly draw connections to human behaviour. We believe that the Incentive Salience hypothesis formulated by Berridge and Robinson \cite{berridge1998role} offers a solution for both of these problems. 
\begin{wrapfigure}{l}{0pt}
        \begin{adjustbox}{width=0.4\columnwidth}
        \tikzset{every picture/.style={line width=0.75pt}} 
        
        \begin{tikzpicture}[x=0.75pt,y=0.75pt,yscale=-1,xscale=1]
        
        \draw   (180.48,130.05) .. controls (180.48,119.06) and (189.39,110.15) .. (200.38,110.15) .. controls (211.37,110.15) and (220.28,119.06) .. (220.28,130.05) .. controls (220.28,141.04) and (211.37,149.95) .. (200.38,149.95) .. controls (189.39,149.95) and (180.48,141.04) .. (180.48,130.05) -- cycle ;
        \draw    (200.38,110.15) .. controls (218.99,91.83) and (257.82,90.88) .. (278.41,108.12) ;
        \draw [shift={(280.57,110.07)}, rotate = 224.02] [fill={rgb, 255:red, 0; green, 0; blue, 0 }  ][line width=0.08]  [draw opacity=0] (10.72,-5.15) -- (0,0) -- (10.72,5.15) -- (7.12,0) -- cycle    ;
        \draw  [dash pattern={on 4.5pt off 4.5pt}]  (280.57,150.02) .. controls (261.36,169.14) and (222.2,169.5) .. (202.45,151.93) ;
        \draw [shift={(200.38,149.95)}, rotate = 405.89] [fill={rgb, 255:red, 0; green, 0; blue, 0 }  ][line width=0.08]  [draw opacity=0] (10.72,-5.15) -- (0,0) -- (10.72,5.15) -- (7.12,0) -- cycle    ;
        \draw   (260.6,130.05) .. controls (260.6,119.01) and (269.54,110.07) .. (280.57,110.07) .. controls (291.6,110.07) and (300.55,119.01) .. (300.55,130.05) .. controls (300.55,141.08) and (291.6,150.02) .. (280.57,150.02) .. controls (269.54,150.02) and (260.6,141.08) .. (260.6,130.05) -- cycle ;
        \draw   (170.47,80.5) -- (310.47,80.5) -- (310.47,180.83) -- (170.47,180.83) -- cycle ;
        
        \draw (200.38,130.05) node  [font=\normalsize]  {$I$};
        \draw (280.57,130.05) node  [font=\normalsize]  {$O^{\pm}$};
        \draw (240.14,105.95) node  [font=\normalsize]  {$b^{\pm }_{t}$};
        \draw (239.81,151.81) node  [font=\normalsize]  {$r^{\pm }_{t}$};
        \draw (190.45,69.01) node  [font=\normalsize,rotate=-0.16]  {$Env_{t}$};
    
    \end{tikzpicture}
    \end{adjustbox}
\caption{\textbf{The process of incentive salience attribution}. Solid and dashed lines represent observable and latent variables.}
\label{fig: incs}
\end{wrapfigure}
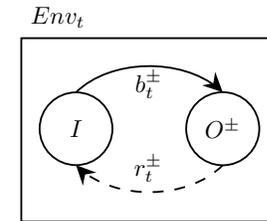
From a behavioural point of view, the Incentive Salience hypothesis states that through repeated interactions (i.e. operant conditioning), objects, $O$, which elicit rewarding experiences, $r$, become valuable, i.e. they acquire salience, to the individual, $I$, interacting with them \cite{berridge1998role, mcclure2003computational}. The amount of salience then controls how likely and intense future interactions between $I$ and $O$ will be \cite{berridge1998role, mcclure2003computational}. Let $(b_t)_{t=1}^{n}$ represent the history of interactions between $I$ and $O$, $(r_t)_{t=1}^{n}$ a measure of how rewarding an interaction $O$ is perceived to be by $I$ and $(Env_t)_{t=1}^{n}$ the changes in the environment in which $I$ and $O$ interact. Following Figure \ref{fig: incs}, we can imagine the intensity of each $b_{t+1}$ increasing and decreasing according to the level of $r_t$. Changes in $r_t$ alter the salience of $O$, which acts as an attracting force for future $b$. Reformulating this in a video-game context, $I$ and $O$ indicate a user and a specific video-game respectively, $b_{t}$ is the amount of playing activity during a gaming session $t$, $r_{t}$ stands for how rewarding the user found the gaming session. Finally $Env$ are all those external factors interfering or promoting the gaming activity (e.g. school days or holidays) which therefore need to be taken into account for estimating an unbiased measure of $r_t$. For convenience, from now on $I$ and $O$ will be used as synonyms for representing a user and a specific game the user is interacting with. The concept of attributed salience will be used interchangeably with that of level of engagement since the two are strongly interconnected and do not differ from a behavioural point of view: high levels of attributed salience pushes $I$ to interact (i.e. engage) more with $O$ \cite{boyle2012engagement,o2008user, berridge1998role}. By keeping this notation we also highlight how our current approach can in theory be extended to contexts other than video games.

\subsection{From Theory to Modelling}
We believe it is of core importance to estimate the level of salience $I$ is attributing to $O$ at a specific $t$. As we have seen before this, other than simply providing a way to assess the current state of the interaction between $I$ and $O$, would allow us to perform informed estimations of all $b_{t+1 : n}$. Following our theoretical framework we hypothesise that, being $f$, $g$ and $k$ a set of unknown arbitrarily complex functions:

\begin{equation}
    b_{t+1 : n} \sim f(O^{\pm})
\label{Eqn1}
\tag{Eqn 1}
\end{equation}

This can be achieved by taking into consideration the full history of rewarding experiences provided to $I$ by $O$:

\begin{equation}
    O^{\pm} \sim g(r_t, r_{t-1}, \dots, r_{t-n})
\label{Eqn2}
\tag{Eqn 2}
\end{equation}

Because $r_t$ is a latent variable it is necessary to find an observable outcome able to approximate it. Again, following our theoretical framework we hypothesize that $r_t$ can be inferred from the full history of observed $b$ between $I$ and $O$ weighted by the effect of $Env$
\begin{equation}
    r_t \sim k((b \cdot Env)_t, (b \cdot Env)_{t-1}, \dots, (b \cdot Env)_{t-n})
\label{Eqn3}
\tag{Eqn 3}
\end{equation}

To summarize, according to our hypothesis the latent variable $r_t$ (which control changes in the salience attributed to $O$) can be inferred from the intensity of observed behaviours produced by $I$ when interacting with $O$.

\subsection{Choosing the Right Modelling Approach}
Following the formulations above we can see how a suitable approach for estimating attributed salience might be to adopt an auto-regressive-like (AR) model with order $p = n$. This model would take as input sequences of $b$, $O$ and $Env$ and attempt to estimate the intensity of all future $b$ generated by $I$ in response to $O$. The model should represent the temporal relationships between the elements in the input sequence. We showed in our previous work \cite{bonometti2019modelling} that simply considering full history of past behaviours without explicitly modelling temporality lead to sub-optimal results. Finally, since our theoretical framework doesn't explicitly indicate which type of function best describes the relationship between the history of interactions and the intensity of future interactions, it is necessary for our modeling approach to be able to learn arbitrary complex functions. We argue therefore that the use of Artificial Neural Network (ANN), and in particular recurrent variants (RNN), are particularly well suited modelling techniques. 

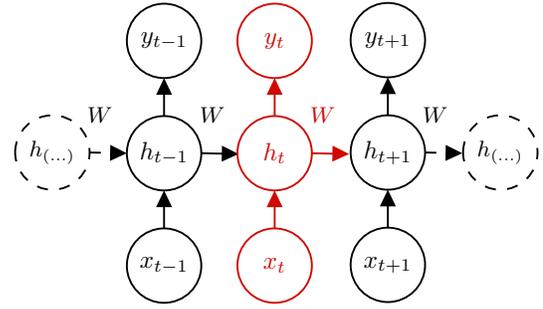
\begin{figure}[h]
    \begin{center}
    \begin{adjustbox}{width=0.8\columnwidth}
        \tikzset{every picture/.style={line width=0.75pt}} 
        
        \begin{tikzpicture}[x=0.75pt,y=0.75pt,yscale=-1,xscale=1]
        
        \draw  [dash pattern={on 4.5pt off 4.5pt}] (121.4,140.06) .. controls (121.4,129.09) and (130.29,120.2) .. (141.26,120.2) .. controls (152.23,120.2) and (161.12,129.09) .. (161.12,140.06) .. controls (161.12,151.03) and (152.23,159.92) .. (141.26,159.92) .. controls (130.29,159.92) and (121.4,151.03) .. (121.4,140.06) -- cycle ;
        \draw   (181,140.46) .. controls (181,129.49) and (189.89,120.6) .. (200.86,120.6) .. controls (211.83,120.6) and (220.72,129.49) .. (220.72,140.46) .. controls (220.72,151.43) and (211.83,160.32) .. (200.86,160.32) .. controls (189.89,160.32) and (181,151.43) .. (181,140.46) -- cycle ;
        \draw   (300.4,140.06) .. controls (300.4,129.09) and (309.29,120.2) .. (320.26,120.2) .. controls (331.23,120.2) and (340.12,129.09) .. (340.12,140.06) .. controls (340.12,151.03) and (331.23,159.92) .. (320.26,159.92) .. controls (309.29,159.92) and (300.4,151.03) .. (300.4,140.06) -- cycle ;
        \draw  [dash pattern={on 4.5pt off 4.5pt}] (360.4,140.06) .. controls (360.4,129.09) and (369.29,120.2) .. (380.26,120.2) .. controls (391.23,120.2) and (400.12,129.09) .. (400.12,140.06) .. controls (400.12,151.03) and (391.23,159.92) .. (380.26,159.92) .. controls (369.29,159.92) and (360.4,151.03) .. (360.4,140.06) -- cycle ;
        \draw   (181,80.46) .. controls (181,69.49) and (189.89,60.6) .. (200.86,60.6) .. controls (211.83,60.6) and (220.72,69.49) .. (220.72,80.46) .. controls (220.72,91.43) and (211.83,100.32) .. (200.86,100.32) .. controls (189.89,100.32) and (181,91.43) .. (181,80.46) -- cycle ;
        \draw   (300.4,80.06) .. controls (300.4,69.09) and (309.29,60.2) .. (320.26,60.2) .. controls (331.23,60.2) and (340.12,69.09) .. (340.12,80.06) .. controls (340.12,91.03) and (331.23,99.92) .. (320.26,99.92) .. controls (309.29,99.92) and (300.4,91.03) .. (300.4,80.06) -- cycle ;
        \draw   (181,200.46) .. controls (181,189.49) and (189.89,180.6) .. (200.86,180.6) .. controls (211.83,180.6) and (220.72,189.49) .. (220.72,200.46) .. controls (220.72,211.43) and (211.83,220.32) .. (200.86,220.32) .. controls (189.89,220.32) and (181,211.43) .. (181,200.46) -- cycle ;
        \draw   (300.4,200.06) .. controls (300.4,189.09) and (309.29,180.2) .. (320.26,180.2) .. controls (331.23,180.2) and (340.12,189.09) .. (340.12,200.06) .. controls (340.12,211.03) and (331.23,219.92) .. (320.26,219.92) .. controls (309.29,219.92) and (300.4,211.03) .. (300.4,200.06) -- cycle ;
        \draw  [dash pattern={on 4.5pt off 4.5pt}]  (161.12,140.06) -- (178,140.4) ;
        \draw [shift={(181,140.46)}, rotate = 181.15] [fill={rgb, 255:red, 0; green, 0; blue, 0 }  ][line width=0.08]  [draw opacity=0] (8.93,-4.29) -- (0,0) -- (8.93,4.29) -- cycle    ;
        \draw    (220.72,140.46) -- (237.6,140.8) ;
        \draw [shift={(240.6,140.86)}, rotate = 181.15] [fill={rgb, 255:red, 0; green, 0; blue, 0 }  ][line width=0.08]  [draw opacity=0] (8.93,-4.29) -- (0,0) -- (8.93,4.29) -- cycle    ;
        \draw  [dash pattern={on 4.5pt off 4.5pt}]  (340.12,140.06) -- (357,140.4) ;
        \draw [shift={(360,140.46)}, rotate = 181.15] [fill={rgb, 255:red, 0; green, 0; blue, 0 }  ][line width=0.08]  [draw opacity=0] (8.93,-4.29) -- (0,0) -- (8.93,4.29) -- cycle    ;
        \draw    (200.86,180.6) -- (200.86,163.32) ;
        \draw [shift={(200.86,160.32)}, rotate = 450] [fill={rgb, 255:red, 0; green, 0; blue, 0 }  ][line width=0.08]  [draw opacity=0] (8.93,-4.29) -- (0,0) -- (8.93,4.29) -- cycle    ;
        \draw    (320.26,180.2) -- (320.26,162.92) ;
        \draw [shift={(320.26,159.92)}, rotate = 450] [fill={rgb, 255:red, 0; green, 0; blue, 0 }  ][line width=0.08]  [draw opacity=0] (8.93,-4.29) -- (0,0) -- (8.93,4.29) -- cycle    ;
        \draw    (200.86,120.6) -- (200.86,103.32) ;
        \draw [shift={(200.86,100.32)}, rotate = 450] [fill={rgb, 255:red, 0; green, 0; blue, 0 }  ][line width=0.08]  [draw opacity=0] (8.93,-4.29) -- (0,0) -- (8.93,4.29) -- cycle    ;
        \draw    (320.26,120.2) -- (320.26,102.92) ;
        \draw [shift={(320.26,99.92)}, rotate = 450] [fill={rgb, 255:red, 0; green, 0; blue, 0 }  ][line width=0.08]  [draw opacity=0] (8.93,-4.29) -- (0,0) -- (8.93,4.29) -- cycle    ;
        \draw  [color={rgb, 255:red, 208; green, 9; blue, 2 }  ,draw opacity=1 ] (240,140.46) .. controls (240,129.49) and (248.89,120.6) .. (259.86,120.6) .. controls (270.83,120.6) and (279.72,129.49) .. (279.72,140.46) .. controls (279.72,151.43) and (270.83,160.32) .. (259.86,160.32) .. controls (248.89,160.32) and (240,151.43) .. (240,140.46) -- cycle ;
        \draw  [color={rgb, 255:red, 208; green, 9; blue, 2 }  ,draw opacity=1 ] (240,80.46) .. controls (240,69.49) and (248.89,60.6) .. (259.86,60.6) .. controls (270.83,60.6) and (279.72,69.49) .. (279.72,80.46) .. controls (279.72,91.43) and (270.83,100.32) .. (259.86,100.32) .. controls (248.89,100.32) and (240,91.43) .. (240,80.46) -- cycle ;
        \draw  [color={rgb, 255:red, 208; green, 9; blue, 2 }  ,draw opacity=1 ] (240,200.46) .. controls (240,189.49) and (248.89,180.6) .. (259.86,180.6) .. controls (270.83,180.6) and (279.72,189.49) .. (279.72,200.46) .. controls (279.72,211.43) and (270.83,220.32) .. (259.86,220.32) .. controls (248.89,220.32) and (240,211.43) .. (240,200.46) -- cycle ;
        \draw [color={rgb, 255:red, 208; green, 9; blue, 2 }  ,draw opacity=1 ]   (279.72,140.46) -- (296.6,140.8) ;
        \draw [shift={(299.6,140.86)}, rotate = 181.15] [fill={rgb, 255:red, 208; green, 9; blue, 2 }  ,fill opacity=1 ][line width=0.08]  [draw opacity=0] (8.93,-4.29) -- (0,0) -- (8.93,4.29) -- cycle    ;
        \draw [color={rgb, 255:red, 208; green, 9; blue, 2 }  ,draw opacity=1 ]   (259.86,180.6) -- (259.86,163.32) ;
        \draw [shift={(259.86,160.32)}, rotate = 450] [fill={rgb, 255:red, 208; green, 9; blue, 2 }  ,fill opacity=1 ][line width=0.08]  [draw opacity=0] (8.93,-4.29) -- (0,0) -- (8.93,4.29) -- cycle    ;
        \draw [color={rgb, 255:red, 208; green, 9; blue, 2 }  ,draw opacity=1 ]   (259.86,120.6) -- (259.86,103.32) ;
        \draw [shift={(259.86,100.32)}, rotate = 450] [fill={rgb, 255:red, 208; green, 9; blue, 2 }  ,fill opacity=1 ][line width=0.08]  [draw opacity=0] (8.93,-4.29) -- (0,0) -- (8.93,4.29) -- cycle    ;
        
        \draw (200.86,200.46) node    {$x_{t-1}$};
        \draw (320.26,200.06) node    {$x_{t+1}$};
        \draw (200.86,140.46) node    {$h_{t-1}$};
        \draw (320.26,140.06) node    {$h_{t+1}$};
        \draw (200.86,80.46) node    {$y_{t-1}$};
        \draw (320.26,80.06) node    {$y_{t+1}$};
        \draw (141.26,140.06) node  [font=\small]  {$h_{( ...)}$};
        \draw (380.26,140.06) node  [font=\small]  {$h_{( ...)}$};
        \draw (166.6,120) node  [font=\small]  {$W$};
        \draw (226.6,120) node  [font=\small]  {$W$};
        \draw (345.6,120) node  [font=\small]  {$W$};
        \draw (259.86,200.46) node  [color={rgb, 255:red, 208; green, 9; blue, 2 }  ,opacity=1 ]  {$x_{t}$};
        \draw (259.86,140.46) node  [color={rgb, 255:red, 208; green, 9; blue, 2 }  ,opacity=1 ]  {$h_{t}$};
        \draw (259.86,80.46) node  [color={rgb, 255:red, 208; green, 9; blue, 2 }  ,opacity=1 ]  {$y_{t}$};
        \draw (285.6,120) node  [font=\small,color={rgb, 255:red, 208; green, 9; blue, 2 }  ,opacity=1 ]  {$W$};

        \end{tikzpicture}
    \end{adjustbox}
    \end{center}

\caption{\textbf{Many To Many Recurrent Neural Network}. Adapted from \cite{lecun2015deep}.
}
\label{fig: rnn}
\end{figure}

When using a RNN for sequence to sequence tasks, the neural network takes as input a sequence of vectors of arbitrary length $(x_t)_{t=1}^{n}$ and as target $(y_t)_{t=1}^{n}$, a sequence of vectors of the same length. A weight matrix $W$, with subsequent non-linear activation functions, is then applied recurrently to each $x_t$ and to a latent variable $h_{t-1}$. This will produce the next $h_t$ that is used for estimating $y_t$ as well as for the subsequent operations. Figure \ref{fig: rnn} demonstrates an example of a `many to many' model. We can see how an RNN satisfies the need for modelling the inputs temporally and in an arbitrarily complex manner. Moreover unlike traditional AR models it is not necessary to specify the order $p$, this is something $W$ dynamically infers from the data, which allows to retain information from varying size sequences in a flexible manner. Following Figure \ref{fig: rnn}, the input $x_t$ is a vector carrying information about $b_t$, $O$ and $Env_t$ while $y_t$ a vector of variables indicative of the intensity of $b_{t+1 : n}$. The latent variable $h_t$ would represent the level of salience $I$ is attributing to $O$ at time $t$. $r_t$ is implicitly computed by the operations performed by $W$ and it reflects in the changes of $h_t$. Given how an RNN computes its latent variables, we can see that at each $t$ the associated $h$ will represent the full history of information that have flowed through the model until then. 


\subsection{Manifold Learning}
Despite the fact that ANNs seem to be a suitable approach for translating our theoretical framework in a machine-learned model, we need to assess if the learned model is able to represent a sensible approximation of the process of salience attribution. We believe that the concept of manifold learning can be used to demonstrate this. ANNs are known to be efficient universal function approximators but it is exceptionally hard to draw insights about the underlying learned function. However, one of the core concepts in deep learning, and machine learning in general, is that the data we observe lies on a manifold: a connected region where each point is surrounded by other extremely similar examples \cite{lecun2015deep}. When training an ANN, we can imagine the operations performed by each layer as learning the coordinates of each input point on a manifold that holds a representation (i.e. an embedding) that is useful for subsequent layers. For example, in a supervised learning context, the last layer of an ANN is usually tasked with performing classification or regression while the layer before that provides the best data transformation for that task. With this in mind we argue that if, through architectural choice, we can enforce an ANN to learn a representation $z$ for which

\begin{equation}
     b_{t+1 : n} \sim f(z)
\label{Eqn4}
\tag{Eqn 4}
\end{equation}

where $f$ is an arbitrarily complex function. This representation $z$ should then be able to place individuals with similar characteristics, with respect to the objective, closer in the embedding space. With this in mind we argue that through the analysis of $z$ it should be possible to inspect these characteristics and compare them with those predicted by the theoretical framework used for designing the model. 


\section{METHODOLOGY}
\subsection{Data}
To conduct our experiments, we gathered data from six games published by our partner company, \textit{Square Enix Ltd.}: \emph{Hitman Go} (hmg), \emph{Hitman Sniper} (hms), \emph{Just Cause 3} (jc3), \emph{Just Cause 4} (jc4), \emph{Life is Strange} (lis), and \emph{Life is Strange: Before the Storm} (lisbf). The data-set contained data from 3,240,000 individuals, evenly distributed across the 6 games, and randomly sampled from all users who played between the games release and January 2020. Different from our previous work, \cite{bonometti2019modelling}, we included a series of metrics representing the $Env$ as well as increased the number of target metrics to provide a better behavioural approximation of engagement. Moreover, in this work no set observation period was required since the modelling approach follows an online strategy: given the history of game sessions for a user the model is trained to perform estimation after each session. It was important, when deciding on which metrics should be adopted, to have a minimal and highly generalizable set of features. This feature selection improved the generalizability and usability of our methodology and allowed us to carry out analytical work on the representation learned by our model: relying on a limited set of of input metrics, selected on the basis of an underlying theoretical framework, makes it easier to generate and interpret human readable visualisations. 

\subsubsection{\textbf{The behavioural and object metric}}
Given a set of $(b_t)_{t=1}^{n}$ between $I$ and $O$ (i.e. game sessions performed by a user within a specific game context), we needed a set of behavioural metrics able to represents the intensity of each $b$. Following the indications provided by our theoretical framework, we decided to employ the same behavioural metrics used in our previous work, see Table \ref{metricsdescription}, as they appear to be optimal candidates for representing the intensity of $b$ in behavioural terms. For representing the $O$ generating $(b_t)_{t=1}^{n}$ we simply retrieved a metric indicating the game context to which the behavioural metric are associated.

\begin{table}[h] \centering
\caption{\textbf{Considered Metrics over Sessions}. Taken from \cite{bonometti2019modelling}}
\label{metricsdescription}
\resizebox{0.88\columnwidth}{!}{
\begin{tabular}{@{}ll@{}}
\toprule
\textbf{Metric}            & \textbf{Description}                   \\ \midrule
{Session Time}         & Overall session duration (minutes)              \\ 
{Play Time}            & Session Time spent actively playing (minutes)    \\ 
{Delta Session}        & Temporal distance  between sessions (minutes)   \\ 
{Activity Index}       & Count of user initiated game-play-related actions. E.g.\\ 
                       & `Talk to NPC' or `Acquire Upgrade' were considered valid\\ 
                       & actions while `Click Menu' or `NPC Attacks You' were not.\\
{Activity Diversity}   & Count of unique voluntarily initiated actions \\ 
\bottomrule
\end{tabular}
}
\end{table}

\subsubsection{\textbf{The environment metrics}}
To represent $(Env_t)_{t=1}^{n}$ we needed a set of metrics which provide a high-level description of the environment in which $I$ and $O$ are interacting. We chose the hour of the day, the day of the week and the day of the year to account for elements like weekends, working hours and bank holidays. We additionally consider a metric indicating the user's broad geographic area to account for regional variations.

\subsubsection{\textbf{The target metrics}}
The target metrics need to be a set of behavioural metrics summarising the intensity of all the $(b_t)_{t+1}^{n}$ after a specific $b_t$. These, following the intuitions from \cite{berridge1998role}, can be used as a measure of attributed salience. We extend the traditional metrics of churn probability ($ch$) and survival time ($st$) to also include survival sessions ($st$) and the time the player is absent from the game between the end of one session and the start of the next ($ab$). The two survival target metrics were calculated following the formula:

\begin{equation}
     survival_t = total\: Pt\: or\: Ps - \sum_{n=1}^{t} Pt\: or\: Pn_n
\label{Eqn5}
\tag{Eqn 5}
\end{equation}

with $Pt$ and $Ps$ being respectively $Played\: Time$ and $Played\: Sessions$. The absence metric is simply the time in minutes between the current session and the previous session. We choose a definition and encoding of the churn variable which is robust to outliers as well being able to represent uncertain cases. Given the two criteria:

\begin{enumerate}[label=(\alph*)]
    \item Completing the game
    \item Being inactive for a period equal or greater to:
        \begin{equation}
            inactivity  = 
            Q_3(x) + 1.5 \cdot IQR(x)
        \label{Eqn6}
        \tag{Eqn 6}
        \end{equation}
\end{enumerate}

where $x$ is a vector of of inter-sessions distances for a specific game, we determined the probability of being in a churning state as follow:
\begin{equation}
    churn\: probability = 
        \begin{cases}
            0.0 & if \, a \\
            1.0 & if \, \neg a \land b \\
            0.5 & otherwise
        \end{cases} 
\label{Eqn7}
\tag{7}
\end{equation} 

Summarizing, we argue that the intensity of all the future interactions between $I$ and $O$ can be expressed through the combinations of the four aforementioned metrics:

\begin{equation}
    b_{t+1 : n} \sim (ch, st_t, ss_t, ab_t)
\label{Eqn8}
\tag{Eqn 8}
\end{equation}

because all the four metrics are good quantifiers of the amount and frequency of future behaviour generated by $I$ in response to $O$.  

\subsubsection{\textbf{Data Preparation}}
For each user in our dataset we retrieve a single feature for the object metric and a temporal series for each other metrics, computing the input and target metrics for each recorded game session. We then split the data-set into training and testing subsets (80 and 20 \% of the original dataset) and rescaled the behavioural metrics using the following formula:

\begin{equation}
    scale(x) =\frac{x - \min(x)} {\max(x) - \min(x)}
\label{Eqn9}
\tag{Eqn 9}
\end{equation}

where $x$ is the feature vector to be re-scaled. To avoid the risk of information leakage $\min$ and $\max$ are computed only on the training set. We then proceeded to numerically encode the variables indicative of $O$ and $Env$ because the categorical encoding employed by our models requires to transform all the unique categories in numerical indices \cite{chollet2015keras}.

\subsection{Hypotheses and Models}
To evaluate the assumptions of our theoretical framework we defined a series of hypotheses which we tested both through the comparison of a set of models as well as by inspecting the embedding space learned by our proposed approach. These hypotheses are: 1) Explicitly modelling temporality in the interactions between $I$ and $O$ results in better performance when estimating behavioural proxies of attributed salience. 2) Models able to learn arbitrarily complex functions will produce less error when estimating behavioural proxies of attributed salience. 3) Our modelling methodology is able to generate a latent representation which reasonably approximates the level of salience an individual has attributed to a game. 4) The encoding of the level of salience provided by the model will reflect increases and decreases in the behavioural metrics used for describing the strength of the interactions between $I$ and $O$. For instance, individuals encoded has having had attributed high salience to $O$ will show a history of more frequent and longer interactions. 5) While the patterns above will appear consistently across game contexts (i.e. the distinct $O$) various profiles will be observable within and between different $O$, because the behavioural manifestation of salience attribution partially depends on the nature of $O$. To test these hypotheses we designed and implemented three models with \textit{ad-hoc} characteristics.

\subsubsection{\textbf{Autoregressive-like Models}}
Two AR models of order $p=1$ were implemented for testing hypotheses 1 and 2. More specifically, we used an ElasticNet regression (TD ENet):

\begin{equation}
    b_{t+1 : n} \sim b_t + Env_t + O
\label{Eqn10}
\tag{Eqn 10}
\end{equation}

and a Multi Layer Perceptron (TD MLP) 

\begin{equation}
    b_{t+1 : n} \sim f (b_t, Env_t, O)
\label{Eqn11}
\tag{Eqn 11}
\end{equation}

We can see that both models work under a Markovian assumption that intensity of future interactions with $O$ is reliant only on the current state of $I$, but TD MLP is not constrained to learn only linear functions.

\subsubsection{\textbf{Melchior Model}}
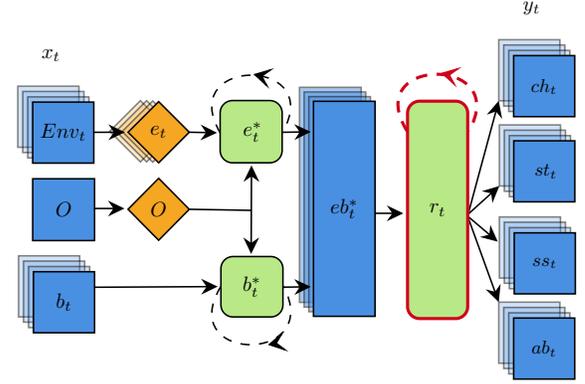
\begin{figure}[h]
\centering
    
    \begin{adjustbox}{width=0.85\columnwidth}
    
        \tikzset{every picture/.style={line width=0.75pt}} 
        
        \begin{tikzpicture}[x=0.75pt,y=0.75pt,yscale=-1,xscale=1]
        
        \draw  [color={rgb, 255:red, 0; green, 0; blue, 0 }  ,draw opacity=0.5 ][fill={rgb, 255:red, 74; green, 144; blue, 226 }  ,fill opacity=0.5 ] (311.2,79.44) -- (350.92,79.44) -- (350.92,218.24) -- (311.2,218.24) -- cycle ;
        \draw  [color={rgb, 255:red, 0; green, 0; blue, 0 }  ,draw opacity=0.5 ][fill={rgb, 255:red, 74; green, 144; blue, 226 }  ,fill opacity=0.5 ] (314.2,82.44) -- (353.92,82.44) -- (353.92,221.24) -- (314.2,221.24) -- cycle ;
        \draw  [color={rgb, 255:red, 0; green, 0; blue, 0 }  ,draw opacity=0.5 ][fill={rgb, 255:red, 74; green, 144; blue, 226 }  ,fill opacity=0.5 ] (317.2,85.44) -- (356.92,85.44) -- (356.92,224.24) -- (317.2,224.24) -- cycle ;
        \draw  [color={rgb, 255:red, 0; green, 0; blue, 0 }  ,draw opacity=0.5 ][fill={rgb, 255:red, 245; green, 166; blue, 35 }  ,fill opacity=0.25 ] (208.55,88.05) -- (228.3,107.8) -- (208.55,127.55) -- (188.8,107.8) -- cycle ;
        \draw  [color={rgb, 255:red, 0; green, 0; blue, 0 }  ,draw opacity=0.5 ][fill={rgb, 255:red, 245; green, 166; blue, 35 }  ,fill opacity=0.25 ] (212.55,88.05) -- (232.3,107.8) -- (212.55,127.55) -- (192.8,107.8) -- cycle ;
        \draw  [color={rgb, 255:red, 0; green, 0; blue, 0 }  ,draw opacity=0.5 ][fill={rgb, 255:red, 245; green, 166; blue, 35 }  ,fill opacity=0.25 ] (216.55,88.25) -- (236.3,108) -- (216.55,127.75) -- (196.8,108) -- cycle ;
        \draw  [color={rgb, 255:red, 0; green, 0; blue, 0 }  ,draw opacity=0.5 ][fill={rgb, 255:red, 74; green, 144; blue, 226 }  ,fill opacity=0.25 ] (129.51,78.71) -- (168.91,78.71) -- (168.91,118.11) -- (129.51,118.11) -- cycle ;
        \draw  [color={rgb, 255:red, 0; green, 0; blue, 0 }  ,draw opacity=0.5 ][fill={rgb, 255:red, 74; green, 144; blue, 226 }  ,fill opacity=0.25 ] (132.71,82.11) -- (172.11,82.11) -- (172.11,121.51) -- (132.71,121.51) -- cycle ;
        \draw  [color={rgb, 255:red, 0; green, 0; blue, 0 }  ,draw opacity=0.5 ][fill={rgb, 255:red, 74; green, 144; blue, 226 }  ,fill opacity=0.25 ] (135.94,85.6) -- (175.34,85.6) -- (175.34,125) -- (135.94,125) -- cycle ;
        \draw  [fill={rgb, 255:red, 74; green, 144; blue, 226 }  ,fill opacity=1 ] (139.1,138.45) -- (179,138.45) -- (179,178.35) -- (139.1,178.35) -- cycle ;
        \draw  [fill={rgb, 255:red, 245; green, 166; blue, 35 }  ,fill opacity=1 ] (220.35,88.65) -- (240.1,108.4) -- (220.35,128.15) -- (200.6,108.4) -- cycle ;
        \draw  [fill={rgb, 255:red, 245; green, 166; blue, 35 }  ,fill opacity=1 ] (220.35,138.15) -- (240.1,158.15) -- (220.35,178.15) -- (200.6,158.15) -- cycle ;
        \draw  [fill={rgb, 255:red, 184; green, 233; blue, 134 }  ,fill opacity=1 ] (260,95.96) .. controls (260,91.54) and (263.58,87.96) .. (268,87.96) -- (292,87.96) .. controls (296.42,87.96) and (300,91.54) .. (300,95.96) -- (300,120.15) .. controls (300,124.57) and (296.42,128.15) .. (292,128.15) -- (268,128.15) .. controls (263.58,128.15) and (260,124.57) .. (260,120.15) -- cycle ;
        \draw  [fill={rgb, 255:red, 184; green, 233; blue, 134 }  ,fill opacity=1 ] (260.5,196.45) .. controls (260.5,192.14) and (263.99,188.65) .. (268.3,188.65) -- (292.7,188.65) .. controls (297.01,188.65) and (300.5,192.14) .. (300.5,196.45) -- (300.5,219.85) .. controls (300.5,224.16) and (297.01,227.65) .. (292.7,227.65) -- (268.3,227.65) .. controls (263.99,227.65) and (260.5,224.16) .. (260.5,219.85) -- cycle ;
        \draw    (178.72,157.64) -- (195.32,157.64) ;
        \draw [shift={(198.32,157.64)}, rotate = 180] [fill={rgb, 255:red, 0; green, 0; blue, 0 }  ][line width=0.08]  [draw opacity=0] (10.72,-5.15) -- (0,0) -- (10.72,5.15) -- (7.12,0) -- cycle    ;
        \draw    (169.6,108.47) -- (195.6,108.41) ;
        \draw [shift={(198.6,108.4)}, rotate = 539.85] [fill={rgb, 255:red, 0; green, 0; blue, 0 }  ][line width=0.08]  [draw opacity=0] (10.72,-5.15) -- (0,0) -- (10.72,5.15) -- (7.12,0) -- cycle    ;
        \draw    (240.1,158.15) -- (280.08,158.84) ;
        \draw    (280.21,132.78) -- (280.01,184) ;
        \draw [shift={(280,187)}, rotate = 270.22] [fill={rgb, 255:red, 0; green, 0; blue, 0 }  ][line width=0.08]  [draw opacity=0] (10.72,-5.15) -- (0,0) -- (10.72,5.15) -- (7.12,0) -- cycle    ;
        \draw [shift={(280.22,129.78)}, rotate = 90.22] [fill={rgb, 255:red, 0; green, 0; blue, 0 }  ][line width=0.08]  [draw opacity=0] (10.72,-5.15) -- (0,0) -- (10.72,5.15) -- (7.12,0) -- cycle    ;
        \draw    (240.1,108.4) -- (255.02,108.7) ;
        \draw [shift={(258.02,108.76)}, rotate = 181.15] [fill={rgb, 255:red, 0; green, 0; blue, 0 }  ][line width=0.08]  [draw opacity=0] (10.72,-5.15) -- (0,0) -- (10.72,5.15) -- (7.12,0) -- cycle    ;
        \draw  [fill={rgb, 255:red, 74; green, 144; blue, 226 }  ,fill opacity=1 ] (139.1,88.95) -- (178.5,88.95) -- (178.5,128.35) -- (139.1,128.35) -- cycle ;
        \draw    (151.1,208.63) -- (255.4,208.06) ;
        \draw [shift={(258.4,208.04)}, rotate = 539.69] [fill={rgb, 255:red, 0; green, 0; blue, 0 }  ][line width=0.08]  [draw opacity=0] (10.72,-5.15) -- (0,0) -- (10.72,5.15) -- (7.12,0) -- cycle    ;
        \draw    (300.2,108.44) -- (315.5,108.2) ;
        \draw [shift={(318.5,108.15)}, rotate = 539.0899999999999] [fill={rgb, 255:red, 0; green, 0; blue, 0 }  ][line width=0.08]  [draw opacity=0] (10.72,-5.15) -- (0,0) -- (10.72,5.15) -- (7.12,0) -- cycle    ;
        \draw  [dash pattern={on 4.5pt off 4.5pt}]  (260.02,108.76) .. controls (231,59.24) and (328.6,59.24) .. (300.2,108.44) ;
        \draw  [dash pattern={on 4.5pt off 4.5pt}]  (260.4,208.04) .. controls (230.4,257.72) and (327.6,258.52) .. (300.8,207.72) ;
        \draw  [fill={rgb, 255:red, 0; green, 0; blue, 0 }  ,fill opacity=1 ] (301.47,248.64) -- (291.77,245.46) -- (299.52,238.83) -- (296.13,244.6) -- cycle ;
        \draw  [fill={rgb, 255:red, 0; green, 0; blue, 0 }  ,fill opacity=1 ] (289.93,78.08) -- (282.4,71.19) -- (292.2,68.34) -- (286.73,72.2) -- cycle ;
        \draw [color={rgb, 255:red, 208; green, 2; blue, 27 }  ,draw opacity=1 ][line width=1.5]  [dash pattern={on 5.63pt off 4.5pt}]  (380.76,107.93) .. controls (349.76,58.93) and (448.71,58.51) .. (420.31,107.71) ;
        \draw  [color={rgb, 255:red, 208; green, 2; blue, 27 }  ,draw opacity=1 ][fill={rgb, 255:red, 208; green, 2; blue, 27 }  ,fill opacity=1 ] (410.93,77.28) -- (403.4,70.39) -- (413.2,67.54) -- (407.73,71.4) -- cycle ;
        \draw  [color={rgb, 255:red, 0; green, 0; blue, 0 }  ,draw opacity=0.5 ][fill={rgb, 255:red, 74; green, 144; blue, 226 }  ,fill opacity=0.25 ] (439.71,48.71) -- (479.11,48.71) -- (479.11,88.11) -- (439.71,88.11) -- cycle ;
        \draw  [color={rgb, 255:red, 0; green, 0; blue, 0 }  ,draw opacity=0.5 ][fill={rgb, 255:red, 74; green, 144; blue, 226 }  ,fill opacity=0.25 ] (442.91,52.11) -- (482.31,52.11) -- (482.31,91.51) -- (442.91,91.51) -- cycle ;
        \draw  [color={rgb, 255:red, 0; green, 0; blue, 0 }  ,draw opacity=0.5 ][fill={rgb, 255:red, 74; green, 144; blue, 226 }  ,fill opacity=0.25 ] (446.14,55.6) -- (485.54,55.6) -- (485.54,95) -- (446.14,95) -- cycle ;
        \draw  [fill={rgb, 255:red, 74; green, 144; blue, 226 }  ,fill opacity=1 ] (449.3,58.95) -- (488.7,58.95) -- (488.7,98.35) -- (449.3,98.35) -- cycle ;
        \draw  [color={rgb, 255:red, 0; green, 0; blue, 0 }  ,draw opacity=0.5 ][fill={rgb, 255:red, 74; green, 144; blue, 226 }  ,fill opacity=0.25 ] (440.11,103.71) -- (479.51,103.71) -- (479.51,143.11) -- (440.11,143.11) -- cycle ;
        \draw  [color={rgb, 255:red, 0; green, 0; blue, 0 }  ,draw opacity=0.5 ][fill={rgb, 255:red, 74; green, 144; blue, 226 }  ,fill opacity=0.25 ] (443.31,107.11) -- (482.71,107.11) -- (482.71,146.51) -- (443.31,146.51) -- cycle ;
        \draw  [color={rgb, 255:red, 0; green, 0; blue, 0 }  ,draw opacity=0.5 ][fill={rgb, 255:red, 74; green, 144; blue, 226 }  ,fill opacity=0.25 ] (446.54,110.6) -- (485.94,110.6) -- (485.94,150) -- (446.54,150) -- cycle ;
        \draw  [fill={rgb, 255:red, 74; green, 144; blue, 226 }  ,fill opacity=1 ] (449.37,113.95) -- (488.77,113.95) -- (488.77,153.35) -- (449.37,153.35) -- cycle ;
        \draw  [color={rgb, 255:red, 0; green, 0; blue, 0 }  ,draw opacity=0.5 ][fill={rgb, 255:red, 74; green, 144; blue, 226 }  ,fill opacity=0.25 ] (439.71,218.11) -- (479.11,218.11) -- (479.11,257.51) -- (439.71,257.51) -- cycle ;
        \draw  [color={rgb, 255:red, 0; green, 0; blue, 0 }  ,draw opacity=0.5 ][fill={rgb, 255:red, 74; green, 144; blue, 226 }  ,fill opacity=0.25 ] (442.91,221.51) -- (482.31,221.51) -- (482.31,260.91) -- (442.91,260.91) -- cycle ;
        \draw  [color={rgb, 255:red, 0; green, 0; blue, 0 }  ,draw opacity=0.5 ][fill={rgb, 255:red, 74; green, 144; blue, 226 }  ,fill opacity=0.25 ] (446.14,225) -- (485.54,225) -- (485.54,264.4) -- (446.14,264.4) -- cycle ;
        \draw  [fill={rgb, 255:red, 74; green, 144; blue, 226 }  ,fill opacity=1 ] (449.3,228.35) -- (488.7,228.35) -- (488.7,267.75) -- (449.3,267.75) -- cycle ;
        \draw  [color={rgb, 255:red, 0; green, 0; blue, 0 }  ,draw opacity=0.5 ][fill={rgb, 255:red, 74; green, 144; blue, 226 }  ,fill opacity=0.25 ] (440.11,163.11) -- (479.51,163.11) -- (479.51,202.51) -- (440.11,202.51) -- cycle ;
        \draw  [color={rgb, 255:red, 0; green, 0; blue, 0 }  ,draw opacity=0.5 ][fill={rgb, 255:red, 74; green, 144; blue, 226 }  ,fill opacity=0.25 ] (443.31,166.51) -- (482.71,166.51) -- (482.71,205.91) -- (443.31,205.91) -- cycle ;
        \draw  [color={rgb, 255:red, 0; green, 0; blue, 0 }  ,draw opacity=0.5 ][fill={rgb, 255:red, 74; green, 144; blue, 226 }  ,fill opacity=0.25 ] (446.54,170) -- (485.94,170) -- (485.94,209.4) -- (446.54,209.4) -- cycle ;
        \draw  [fill={rgb, 255:red, 74; green, 144; blue, 226 }  ,fill opacity=1 ] (449.7,173.35) -- (489.1,173.35) -- (489.1,212.75) -- (449.7,212.75) -- cycle ;
        \draw    (350.3,160.72) -- (375.17,160.75) ;
        \draw [shift={(378.17,160.76)}, rotate = 180.07] [fill={rgb, 255:red, 0; green, 0; blue, 0 }  ][line width=0.08]  [draw opacity=0] (10.72,-5.15) -- (0,0) -- (10.72,5.15) -- (7.12,0) -- cycle    ;
        \draw    (419.6,162.55) -- (438.93,91.01) ;
        \draw [shift={(439.71,88.11)}, rotate = 465.12] [fill={rgb, 255:red, 0; green, 0; blue, 0 }  ][line width=0.08]  [draw opacity=0] (10.72,-5.15) -- (0,0) -- (10.72,5.15) -- (7.12,0) -- cycle    ;
        \draw    (419.6,162.55) -- (438.69,215.29) ;
        \draw [shift={(439.71,218.11)}, rotate = 250.1] [fill={rgb, 255:red, 0; green, 0; blue, 0 }  ][line width=0.08]  [draw opacity=0] (10.72,-5.15) -- (0,0) -- (10.72,5.15) -- (7.12,0) -- cycle    ;
        \draw    (416.32,164.6) -- (437.88,145.12) ;
        \draw [shift={(440.11,143.11)}, rotate = 497.91] [fill={rgb, 255:red, 0; green, 0; blue, 0 }  ][line width=0.08]  [draw opacity=0] (10.72,-5.15) -- (0,0) -- (10.72,5.15) -- (7.12,0) -- cycle    ;
        \draw    (419.6,162.55) -- (436.81,177.11) ;
        \draw [shift={(439.1,179.05)}, rotate = 220.24] [fill={rgb, 255:red, 0; green, 0; blue, 0 }  ][line width=0.08]  [draw opacity=0] (10.72,-5.15) -- (0,0) -- (10.72,5.15) -- (7.12,0) -- cycle    ;
        \draw  [color={rgb, 255:red, 0; green, 0; blue, 0 }  ,draw opacity=0.5 ][fill={rgb, 255:red, 74; green, 144; blue, 226 }  ,fill opacity=0.25 ] (130.01,188.21) -- (169.41,188.21) -- (169.41,227.61) -- (130.01,227.61) -- cycle ;
        \draw  [color={rgb, 255:red, 0; green, 0; blue, 0 }  ,draw opacity=0.5 ][fill={rgb, 255:red, 74; green, 144; blue, 226 }  ,fill opacity=0.25 ] (133.21,191.61) -- (172.61,191.61) -- (172.61,231.01) -- (133.21,231.01) -- cycle ;
        \draw  [color={rgb, 255:red, 0; green, 0; blue, 0 }  ,draw opacity=0.5 ][fill={rgb, 255:red, 74; green, 144; blue, 226 }  ,fill opacity=0.25 ] (136.44,195.1) -- (175.84,195.1) -- (175.84,234.5) -- (136.44,234.5) -- cycle ;
        \draw  [fill={rgb, 255:red, 74; green, 144; blue, 226 }  ,fill opacity=1 ] (139.6,198.45) -- (179,198.45) -- (179,237.85) -- (139.6,237.85) -- cycle ;
        \draw  [fill={rgb, 255:red, 74; green, 144; blue, 226 }  ,fill opacity=1 ] (320.2,88.44) -- (359.92,88.44) -- (359.92,227.24) -- (320.2,227.24) -- cycle ;
        \draw    (300.2,208.44) -- (315.5,208.2) ;
        \draw [shift={(318.5,208.15)}, rotate = 539.0899999999999] [fill={rgb, 255:red, 0; green, 0; blue, 0 }  ][line width=0.08]  [draw opacity=0] (10.72,-5.15) -- (0,0) -- (10.72,5.15) -- (7.12,0) -- cycle    ;
        \draw  [color={rgb, 255:red, 208; green, 2; blue, 27 }  ,draw opacity=1 ][fill={rgb, 255:red, 184; green, 233; blue, 134 }  ,fill opacity=1 ][line width=1.5]  (380.88,96.12) .. controls (380.88,91.83) and (384.35,88.36) .. (388.64,88.36) -- (411.92,88.36) .. controls (416.21,88.36) and (419.68,91.83) .. (419.68,96.12) -- (419.68,220.92) .. controls (419.68,225.21) and (416.21,228.68) .. (411.92,228.68) -- (388.64,228.68) .. controls (384.35,228.68) and (380.88,225.21) .. (380.88,220.92) -- cycle ;
        
        \draw (469,78.65) node  [font=\small]  {$ch_{t}$};
        \draw (470.07,133.65) node  [font=\small]  {$st_{t}$};
        \draw (469.4,193.05) node  [font=\small]  {$ss_{t}$};
        \draw (469,248.05) node  [font=\small]  {$ab_{t}$};
        \draw (158.8,108.65) node  [font=\normalsize]  {$Env_{t}$};
        \draw (159.05,158.4) node  [font=\normalsize]  {$O$};
        \draw (159.3,218.15) node  [font=\normalsize]  {$b_{t}$};
        \draw (400.28,158.52) node  [font=\normalsize]  {$r_{t}$};
        \draw (220.35,108.4) node  [font=\normalsize]  {$e_{t}$};
        \draw (281,108.05) node  [font=\normalsize]  {$e^{*}_{t}$};
        \draw (220.35,158.15) node  [font=\normalsize]  {$O$};
        \draw (280.5,208.15) node  [font=\normalsize]  {$b^{*}_{t}$};
        \draw (461,28) node    {$y_{t}$};
        \draw (151,59) node    {$x_{t}$};
        \draw (340.06,157.84) node  [font=\normalsize]  {$eb^{*}_{t}$};

        \end{tikzpicture}

    \end{adjustbox}
  
\caption{\textbf{Melchior Model Core Architecture.} Blue squares represents inputs, outputs and fully connected operations. Yellow diamonds indicates categorical embedding operations. Green rounded rectangles are recurrent operations. Shapes with shaded copies represents time series. The red highlight indicates, according to what we specified in \ref{Eqn4}, the portion of the model that we hypothesize holding a representation of the salience $I$ is attributing to $O$.}
\label{fig: melchior}
\end{figure}

Additionally, we propose the novel `Melchior Model' (MM), which is specifically designed to implement the insights from our theoretical framework and to test all the aforementioned hypotheses, Figure \ref{fig: melchior}. Our MM architecture models the contribution of each component, $O$, $b$, and $Env$, separately in a way which resembles the specifications of section III.D, each component is then pooled and used for estimating metrics representative of the intensity of future interactions between $I$ and $O$. The guiding concept was multitask learning \cite{lecun2015deep}: given a set of targets and the assumption that these share a common representation, explicitly modelling this last one allows to capture a collection of common factors shared among all targets. We hypothesised that this would have provided two benefits. Firstly, better generalization and superior performance in terms of goodness of fit. Secondly, the ability to inspect an overarching representation able to explain the variance in the target metrics. This representation would constitute an approximation of the concept of attributed salience expressed in section III.B.

All three models take as inputs two series of vectors, $(b_t)_{t=1}^{n}$ and $(Env_t)_{t=1}^{n}$, and a single feature $O$ and are trained to output four series of vectors corresponding to the target metrics. They were trained using Binary Cross Entropy for churn probability and the Symmetric Mean Absolute Percentage Error (SMAPE) for the other metrics. Both loss functions are bounded between 0 and 1 and lower values correspond to better performance. These loss functions were also used for computing the performance of the models on the test set\footref{constrains}.

\subsection{Experiments}

\subsubsection{\textbf{Experiment 1 - Model Comparison}}
The first experiment is designed to replicate the findings of our previous work \cite{bonometti2019modelling} as well as test the assumption made by our theoretical framework that engagement needs to be modelled as a dynamic system. To achieve this we implemented the three models discussed above and compared their performance. We employ 20\% of the training set to search for the best hyper-parameters while keeping the architecture of the three models fixed (i.e. only elements such as the number of layers and hidden units were tuned). This was achieved using the Hyperband algorithm \cite{li2017hyperband} due to its capacity to converge to a good solution with relatively limited computational resources. After the tuning process we fit each model on the entirety of the training set and then compute the evaluation metrics on the test set.

\subsubsection{\textbf{Experiment 2 - Embedding Visualization}}
The second experiment aims to verify, through visual inspection, that MM learns a separate representation for each $O$ that the model observes and that these representations reasonably encode the different level of salience that various $I$ attribute to $O$. To extract these representations we simply truncate the model's weights up to the point highlighted in Figure \ref{fig: melchior} and then transform the testing set through the usual forward pass. For generating the visualization necessary to test our hypothesis we needed a methodology able to represent in two dimensions the type of spatial properties discussed in section III.E  For this reason we employed the Uniform Manifold Approximation and Projection for Dimension Reduction (UMAP) technique \cite{2018arXivUMAP}. UMAP is a manifold learning technique able to represent the global and local geometry of a set of data in a arbitrary number of dimensions, often 2 or 3 for visualization purposes. In general terms this means that, on a two dimensional plane, points are placed closer or further away to each other depending on their similarity (i.e. proximity) in the original space.

\subsubsection{\textbf{Experiment 3 - Embedding Partitioning}}
The final experiment analyzes the applicability of using the representation from experiment two to validate the hypothesis that behavioural traces that are coherent with the incentive salience framework can be extracted from the learned embedding. Following the manifold hypothesis we know that the learned representation should place points similar to each others close in the multidimensional space. Therefore partitioning this space, with distance-based techniques, should allow us to individuate different groups encoded with different levels of attributed salience. To achieve this we run a mini-batch k-means algorithm \cite{sculley2010web, scikit-learn} on the representation learned by the model for each $O$ and then inspect the behavioural metrics associated to each partition. Since k-means requires a specific number of partitions, we used the elbow method for choosing the number of partitions that maximize the marginal gain in explained variance \footref{constrains}. We decided to use the k-means algorithm despite its known limitations \cite{scikit-learn} because of its ability to scale well when $N \geq 10^5$. After performing the partitioning we visualized the temporal traces of the behavioural input for each partition and compared them with the distribution of target metrics for each partition. The code for all the experiments has been written in Python 3.6 relying on Keras \cite{chollet2015keras} for the implementation of the models and on Scikit-learn \cite{scikit-learn} for the partitioning algorithm.


\section{RESULTS}

\subsubsection{\textbf{Experiment 1 - Model Comparison}}
\begin{figure}[h]
\centering
\includegraphics[width=0.85\columnwidth]{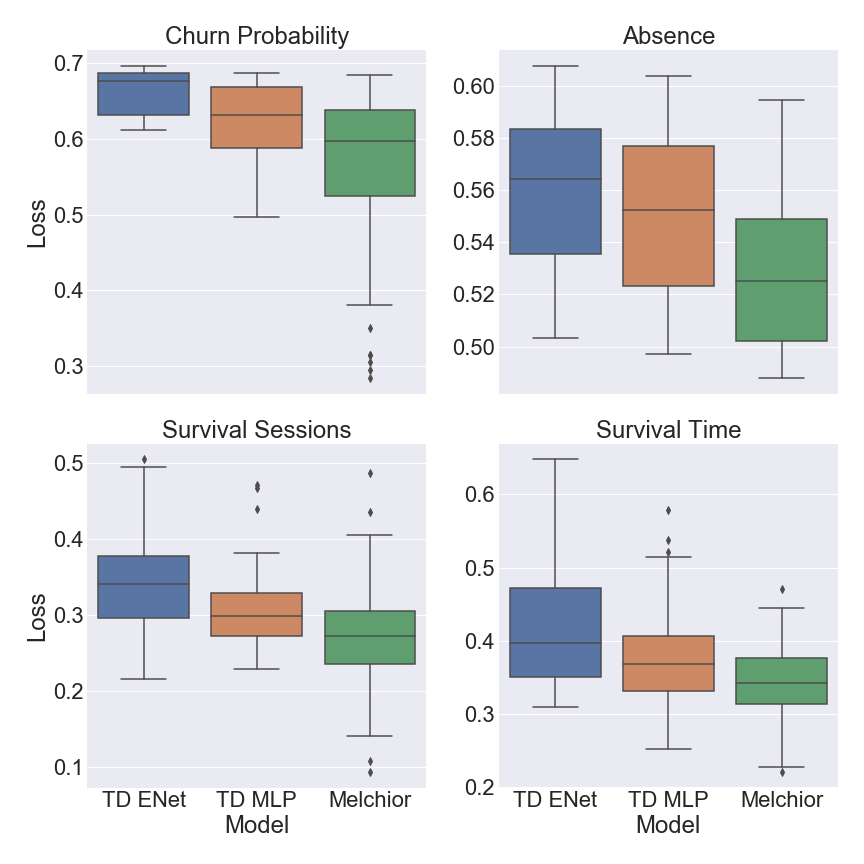}
\caption{\textbf{Models Comparison} The four plots report the performance of the three models}
\label{model_comp}
\end{figure}

Figure \ref{model_comp} shows the performance of each model collapsed over game context and play sessions. Each plot representing the performance for a specific objective. We can see how the performance of the three models follows the expected trend, TD ENet $<$ TD MLP $<$ Melchoir, in all four targets metrics. Additonally, performing this comparison on non-collapsed data \footnote{Omitted due to space concerns. Can be found at https://github.com/vb690/modelling\_engagement\_ammount/wiki \label{constrains}} demonstrates a near-identical trend, strengthening the assumption that temporally modelling interactions is vital for modelling salience attribution.

\subsubsection{\textbf{Experiment 2 - Embedding Visualization}}

\begin{figure}[h]
\centering
\includegraphics[width=0.85\columnwidth]{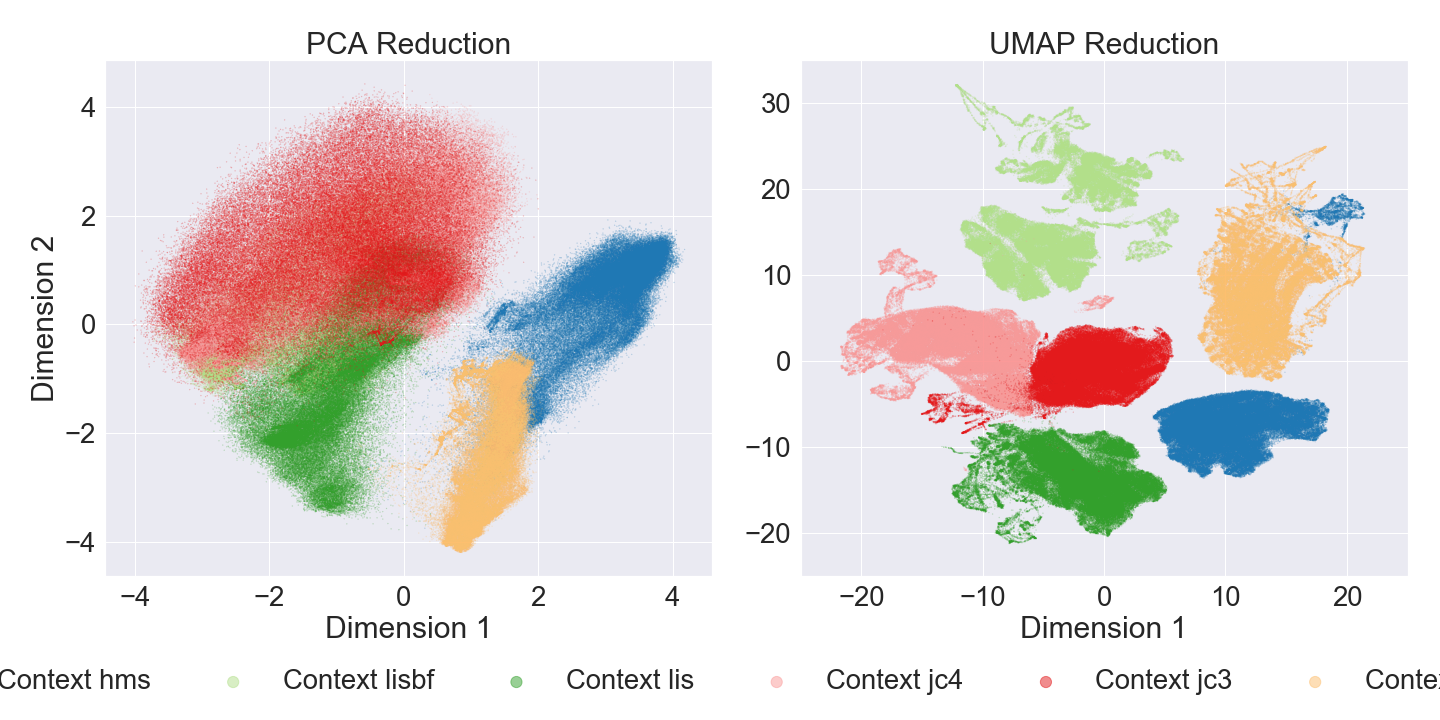}
\caption{\textbf{Learned representation for the six games contexts.} }
\label{context_emb}
\end{figure}

Figure \ref{context_emb} reports the representation learned by MM on the test set. 

\begin{figure}[h]
\centering
\includegraphics[width=0.95\columnwidth]{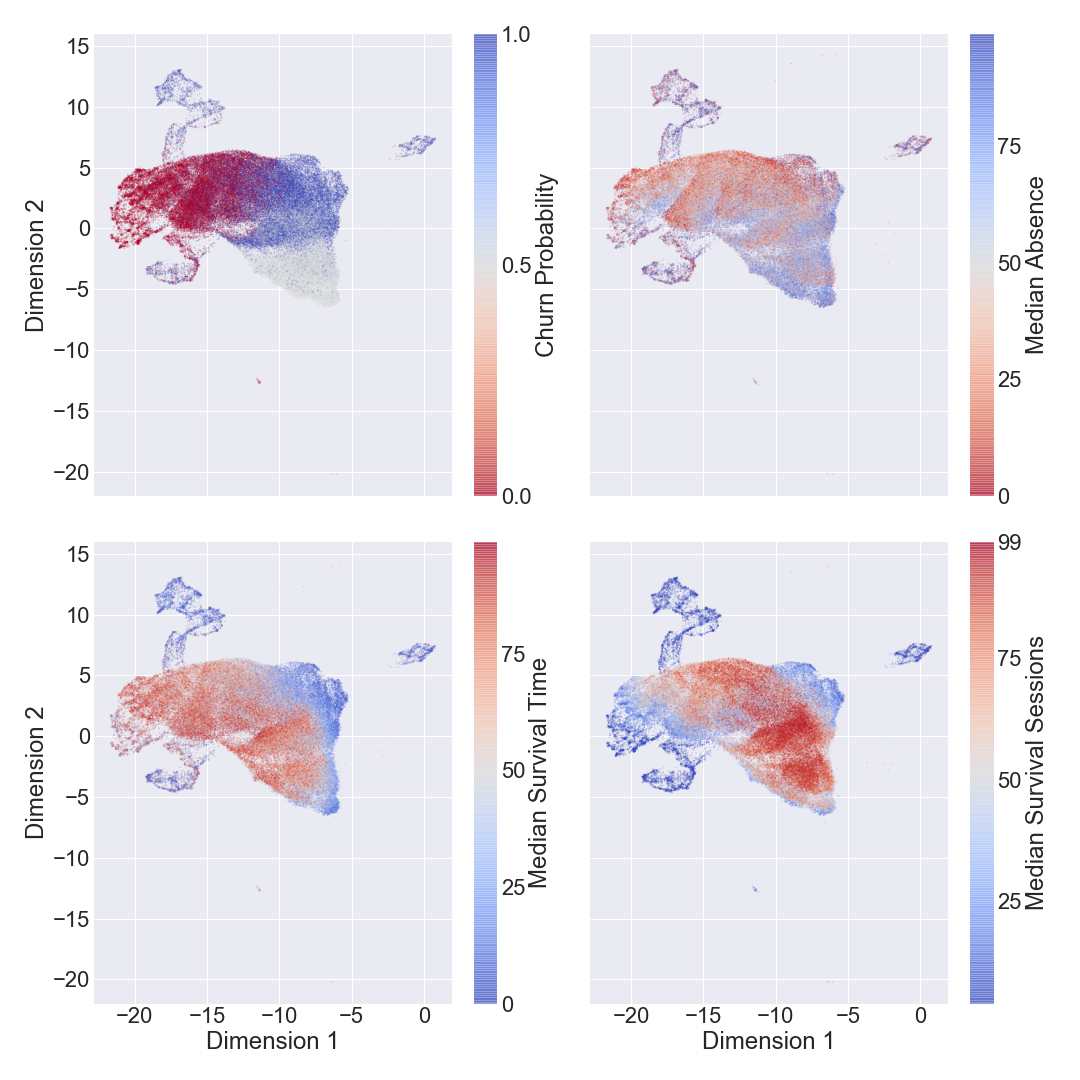}
\caption{\textbf{Visualization of the learned representation for the jc4 context.} Each plot report the UMAP reduction of the learned representation colour coded according to the median value of the four target metrics. From a behavioural point of view, blue colors might be interpreted as encoding low attributed salience while red colors high attributed salience. We advise to consult the ancillary results for examining the entire space.}
\label{targets_emb}
\end{figure}

We see that UMAP is able to more faithfully represents global geometry, placing points from different games within distinct regions, compared to traditional Principal Component Analysis (PCA). Focusing on jc4 and knowing that UMAP attempts to preserve the local geometry, we can see from Figure \ref{targets_emb} that MM was able to generate a representation where users with similar behavioural measures of attributed salience (i.e. the four target metrics) are placed closer to each other in the embedding space. The fact that each metric is represented over different and partially non-overlapping areas of the embedding space indicates that the model learned the type of general representation that we illustrated with \ref{Eqn4} and that is enforced by multitask learning. Such representation can be considered as an abstract and overarching factor, much like the concepts of engagement and attributed salience, able to explain the variations in the behavioural targets.

\subsubsection{\textbf{Experiment 3 - Embedding Partitioning}}
\begin{figure}[h]
\centering
\includegraphics[width=0.70\columnwidth]{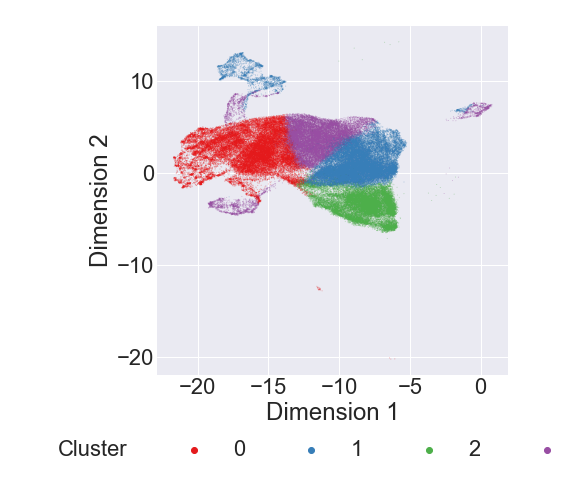}
\caption{\textbf{Visualization of the individuated partitions for the jc4 context.} }
\label{cluster_emb}
\end{figure}

\begin{figure*}[h]
\centering
\includegraphics[width=0.85\textwidth]{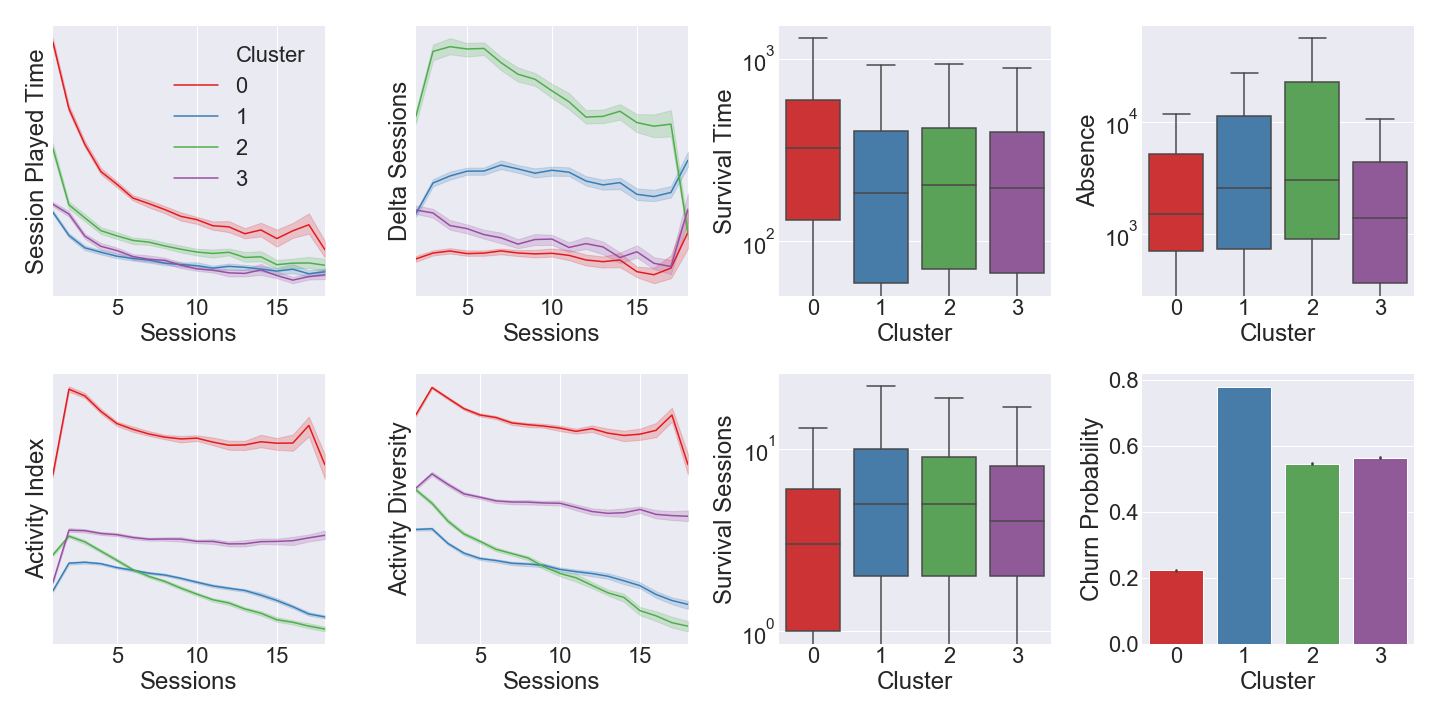}
\caption{\textbf{Characteristics of the individuated partitions.} The line-plots on the left show the progression over time of the input metrics, with the solid line indicating the average and the shaded area the 95\% confidence interval. The plots on the right indicates the distribution of the target metrics.}
\label{cluster_profile}
\end{figure*}

Figure \ref{cluster_emb} shows the results of K-Means partitioning on the embedding space. Looking at the characteristics of the partitions in Figure \ref{cluster_profile} we can see how each target metric develops differently over time but that a general pattern seems to control each partition's trace. Partitions which have a low measurement in the target metrics representing attributed salience are shown to have shorter and less frequent interactions with the game, while the opposite appears to be the case for those partitions which have a high measurement in the same target metrics\footnote{Consulting the ancillary results we can see that this pattern is consistent through all the games.}. 

\section{DISCUSSION AND LIMITATIONS}

\subsubsection{\textbf{Theoretical Implications}}
Our theoretical framework indicates that, from a behavioural point of view, $I$ who attribute a high level of salience to a specific $O$ will show a history of frequent and long interactions with $O$. Following the formulation by \cite{o2008user} this would mean that $I$ stays in the sustained engagement state for longer, or frequently re-engage after periods of disengagement. According to our target metrics this would mean not churning, being absent for shorter and "surviving" for longer periods of time. Focusing on cluster number zero in figure \ref{cluster_profile} we see that it is the cluster that according to the distribution of its target metrics could be hypothesized having the highest level of attributed salience to playing the specific game. If we then look at the behavioural traces describing the interactions between the $I$ in this cluster and the game, we can see that they perfectly match with what is prescribed by our theoretical framework (see sections III.a and III.b). We see that the same logic applies for all the other partitions but with slight differences indicative of the different manifestations of the engagement process. This, in addition to the results provided by experiment 1 and 2, shows that the theoretical constraints imposed on our model do not just lead to better performance when compared with unconstrained methodologies but also forces the model to learn a representations that appear to be coherent with our theoretical assumptions.

\subsubsection{\textbf{Practical Implications}}
As a byproduct of the present work we produced a scalable cross-game model for estimating metrics that are of core importance to industry applications (i.e. frequency and amount of future playing behaviour). The same model can be used as a feature extractor given its capacity to learn representations that are reasonable approximations of the quality of the interaction between a user and a game. Finally, the methodology we used for analyzing the learned embedding offers an efficient way to perform time series clustering, a task often crippled by prohibitive time and memory demands. This is done by first constructing  a compact and static representation of the history of interactions between the user and the game and subsequently partitioning it. Moreover, the learned representation, when compared with the raw features traditionally used in the literature, has the advantage of having already been transformed by the network to best describe the target metrics.

\subsubsection{\textbf{Limitations and Future Work}}
This work has a series of limitations. 1) The analysis conducted for verifying the linkage between theory, model and observed behaviours is still preliminary and more careful and precise investigations would likely yield insights. 2) Due to the assumptions that k-means makes about the shape of the clusters it may be the case that more sophisticated approaches provide a higher-quality space partitioning. 3) For verifying that our modelling approach can be extended to general human behaviour we have to preform the same analysis done in this work with data coming from $O$ that are not limited to video games. 4) Our approach appears to be suitable for modelling the amount of engagement or attributed salience but we can't say anything about the factors of $I$, $O$ and $Env$ which control the changes in the these constructs. 5) Differently to \cite{reguera2020quantifying, bauckhage2012players} our approach doesn't explicitly produce a clear mathematical formulation explaining the observed changes in behaviour. This can at most be inferred or hypothesized observing the behavioral traces in Figure \ref{cluster_profile}. All these constraints constitute venues for future works.


\bibliographystyle{IEEEtran}
\bibliography{main}

\end{document}